\documentclass[conference]{IEEEtran}
\IEEEoverridecommandlockouts

\usepackage{cite}
\usepackage{amsmath,amssymb,amsfonts}
\usepackage{algorithmic}
\usepackage{graphicx}
\usepackage{textcomp}
\usepackage{xcolor}
\usepackage{hyperref}
\def\BibTeX{{\rm B\kern-.05em{\sc i\kern-.025em b}\kern-.08em
    T\kern-.1667em\lower.7ex\hbox{E}\kern-.125emX}}
\begin{document}

\title{Composite Data Augmentations for Synthetic Image Detection against Real-World Perturbations}



\author{
\IEEEauthorblockN{Efthymia Amarantidou$^1$, Christos Koutlis$^2$, Symeon Papadopoulos$^2$, Panagiotis C. Petrantonakis$^1$}
\IEEEauthorblockA{\textit{$^1$Aristotle University of Thessaloniki, Greece} \\
\{efthamar,ppetrant\}@ece.auth.gr}
\IEEEauthorblockA{\textit{$^2$Centre for Research and Technology Hellas, Greece} \\
\{ckoutlis,papadop\}@iti.gr}
}


\maketitle

\begin{abstract}
The advent of accessible Generative AI tools enables anyone to create and spread synthetic images on social media, often with the intention to mislead, thus posing a significant threat to online information integrity. Most existing Synthetic Image Detection (SID) solutions struggle on generated images sourced from the Internet, as these are often altered by compression and other operations. To address this, our research enhances SID by exploring data augmentation combinations, leveraging a genetic algorithm for optimal augmentation selection, and introducing a dual-criteria optimization approach. These methods significantly improve model performance under real-world perturbations. Our findings provide valuable insights for developing detection models capable of identifying synthetic images across varying qualities and transformations, with the best-performing model achieving a mean average precision increase of +22.53\% compared to models without augmentations.  The implementation is available at \href{https://github.com/efthimia145/sid-composite-data-augmentation}{github.com/efthimia145/sid-composite-data-augmentation}.
\end{abstract}


\section{Introduction}
The proliferation of generative models, such as Generative Adversarial Networks (GANs), Variational Autoencoders (VAEs) and Diffusion Models (DMs) has revolutionized fields like art, content creation, business and scientific research. However, these advancements, particularly in image generation technology, also pose significant challenges and increase the necessity of being able to distinguish synthetic images from real ones. The widespread dissemination of synthetic images on social media platforms, often subjected to post-processing operations- like compression and cropping-  as well as potential additional modifications like blurring and noise, make their detection a hard challenge. 

This paper addresses the SID task and proposes a set of methodologies aimed at improving model performance against common image post-processing operations. Our primary contributions include:
\begin{itemize}
    \item \textbf{Investigation of optimal augmentation combinations} for training SID models, demonstrating gains of +22.53\% compared to the model trained without augmentations.
    \item \textbf{Utilization of genetic algorithm for augmentation selection}, 
     exploring a structured approach to examine the wide augmentations space and identify strategies for optimal model performance against real-world perturbations.
    \item \textbf{A dual-criteria optimization approach}, incorporating classification loss and feature similarity loss, achieving comparable mAP and increased accuracy of 1.39\%–4.08\% over the model trained solely with the optimal augmentation combination.
    \item \textbf{Comprehensive experimental analysis} evaluating our methods' effectiveness in scenarios with perturbations.
\end{itemize}
By addressing these challenges, this research contributes to the development of SID models that sustain high performance in the presence of real-world perturbations. The findings presented provide valuable insights for future advancements.

\section{Related Work}
Advancements in generative models, such as GANs and diffusion models, have significantly increased the realism of synthetic images, driving research on enhancing synthetic image detectors. Early work by Chai et al. \cite{b1} identifies persistent detectable artifacts across images from different generative methods, while Wang et al. \cite{b2} focus on the creation of a  universal detector that generalizes across architectures and datasets. Ojha et al. \cite{b3} utilize the feature space of a large pretrained vision-language model for universal fake detection, while Liu et al.\cite{b4} introduce FatFormer, a transformer-based method that integrates local forgery traces within image and frequency domains and enhances generalization through language-guided alignment with image and text prompts. Koutlis and Papadopoulos \cite{b5} leverage intermediate blocks of CLIP's \cite{b6} image encoder and map them  to a learnable forgery-aware vector space, improving generalization. 

\paragraph{Robustness and Detection under Perturbations}
Recent research focuses on models' performance under perturbations. Corvi et al. \cite{b7} analyze diffusion-generated images and test their methods both on resized and compressed data, observing a general reduction of the performance.  Wang et al. \cite{b2} show that data augmentation enhances robustness against post-processing, such as compression, blurring and resizing and highlight that augmentation improves the model performance under perturbations. Building on this framework, Gragnaniello et al. \cite{b8} introduce variations and apply stronger augmentations (Gaussian noise, geometric transformations, cut-out, brightness/contrast changes) during the training process, evaluating under different compression levels and resizing factors. Recent studies like \cite{b9}, \cite{b10}, \cite{b11} demonstrate that leveraging large vision foundation models like CLIP \cite{b6} enhances the training of universal detectors, resulting in classifiers with strong generalization and robustness to perturbations. Purnekar et al. \cite{b24} introduce the print-and-scan (P\&S) augmentation to force detectors to learn more robust features that can be detected even after other forms of processing.

\paragraph{Evaluation Methods}
Synthetic image detection models are evaluated using accuracy, precision-recall, and mean average precision (mAP), with performance tested under perturbations like JPEG compression, blurring, noise and resizing. ForenSynths \cite{b2}, a benchmark dataset designed to assess detection generalization across generative models, plays a key role in evaluation.
SIDBench \cite{b12} further standardizes assessments by integrating multiple detectors and testing their performance under real-world transformations.


\section{Methodology}
Our work builds on \cite{b2} by employing a structured augmentation selection process. Our methodology involves three main approaches: Augmentation-enhanced training with a greedy approach, Augmentation-enhanced training with a Genetic Algorithm and training with dual-criteria optimization.

    \subsection{Augmentation-Enhanced Training}
    
\begin{figure}[htbp]
    \centerline{\includegraphics[clip, trim=0cm 3cm 0cm 1cm, width=0.4\textwidth]{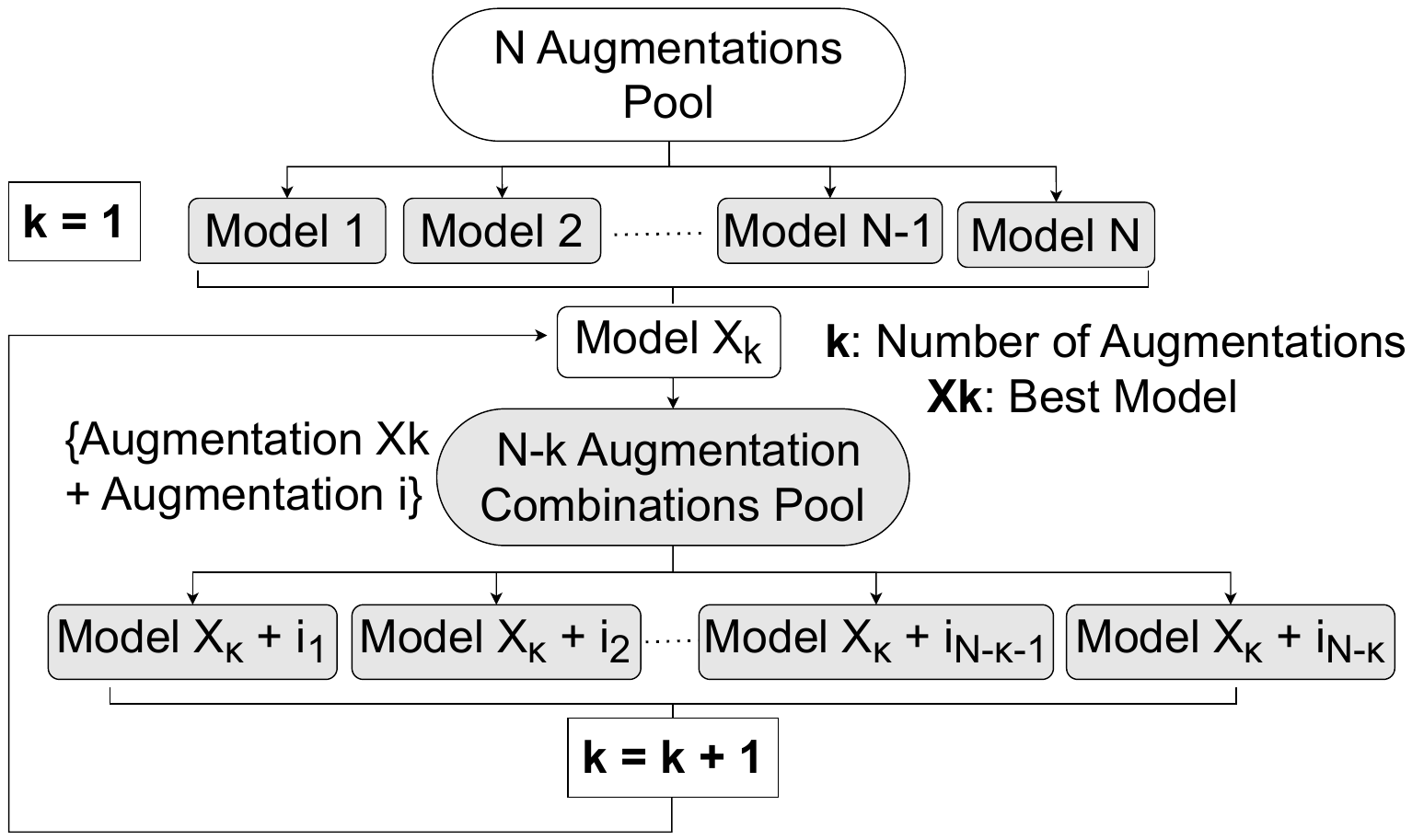}}
    \caption{Greedy Approach for the selection of optimal augmentations}
    \label{fig:greedy-method}
    \end{figure}

    \paragraph{Greedy Approach}
    We follow a greedy search strategy to identify the best combination of image augmentation techniques to use during the training process that optimize model performance. An initial set of $N$ augmentations is considered,  $N$ models are trained individually with each augmentation and their performance is evaluated. The best performing augmentation is selected as the base. Then, $N-1$ models are trained by sequentially combining the selected augmentation with others. This iterative process continues until an optimal augmentation set is determined based on performance metrics. The process is demonstrated in Fig. \ref{fig:greedy-method}.

    \paragraph{Genetic Algorithm}
    Since our approach for improving model performance in real-world perturbation scenarios is based on augmentation-enhanced model training, the search space of possible augmentation combinations is very large. To this end, we employ a genetic algorithm (GA) as a structured approach to efficiently explore this space and identify the optimal augmentations. An initial population with randomly selected augmentation combinations from the original predefined pool is selected. Each individual represents a potential combination where each augmentation is encoded as a binary value (1 for enabled, 0 for disabled). The candidates are evaluated based on model performance on the most demanding evaluation scenario explained later on. Selection is performed using a tournament strategy, followed by crossover and mutation to generate new candidates. The process iterates for a predefined number of generations, with the best-performing augmentations selected as the final solution.

    \subsection{Training with dual-criteria optimization}
\begin{figure}[htbp]
    \centerline{\includegraphics[clip, trim=0cm 6.5cm 1.8cm 4.2cm, width=0.4\textwidth]{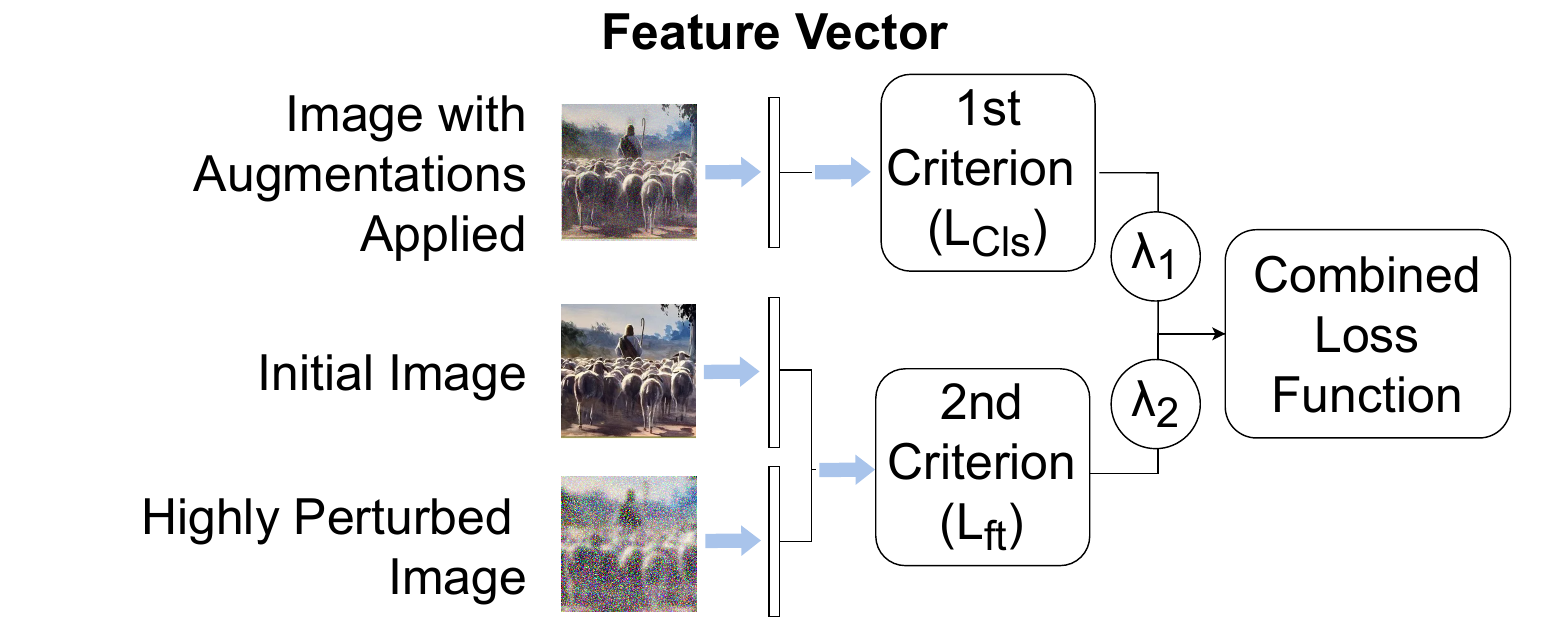}}
    \caption{Methodology for the dual-criteria optimization approach}
    \label{fig:dual-criteria-method}
    \end{figure}

    We also experimented with a more structured learning approach and implemented a dual-criteria optimization strategy. The first criterion is a Binary Cross Entropy (BCE) Loss and is focused on classification accuracy ($L_{cls}$), ensuring precise differentiation between synthetic and real images. For the second criterion, we experimented with a Mean Squared Error (MSE) and a Cosine Similarity Loss function, to minimize feature variations between perturbed and unperturbed versions of the same image ($L_{ft}$), enforcing invariance to distortions. By jointly optimizing these objectives as defined in \eqref{eq:double-criteria-loss}, the goal is for the model to learn perturbations-invariant features. 

    \begin{equation}\label{eq:double-criteria-loss}
    L_{tot} = \lambda_{1} L_{cls} + \lambda_{2} L_{ft},
    \end{equation}
where $\lambda_{1}$, $\lambda_{2}$ define the weight of each loss function. The process is demonstrated in Fig. \ref{fig:dual-criteria-method}.

    \subsection{Model Evaluation}
    
    To assess the effectiveness of the proposed training strategies, a comprehensive evaluation framework is employed. Models are tested across different conditions, including unaltered images, images with individual perturbations and images with combined perturbations applied. The evaluation is conducted on images from multiple datasets following the protocol introduced in \cite{b2}. Performance is quantified using metrics such as Average Precision (AP), Accuracy, and mean Average Precision (mAP) across all evaluation datasets. The mean Average Precision (mAP) is defined in \eqref{eq:map}. Additionally, the improvement introduced by augmentations is measured through mAP Gain, defined in \eqref{eq:map-gain}, which compares the mAP of models trained with the use of augmentations to a baseline model trained without augmentations.

    \begin{equation}\label{eq:map}
            mAP = \frac{1}{D} \sum_{d=1}^{D} AP_{d},
    \end{equation}
where:
    \begin{itemize}
        \item $D$: the number of total evaluation datasets
        \item $AP_{d}$: the Average Precision in dataset $d$
    \end{itemize}

    \begin{equation}\label{eq:map-gain}
        \text{mAP Gain} = \frac{mAP\text{-} mAP_{baseline}}{mAP_{baseline}} \times 100\%
    \end{equation}

\section{Experimental Design}

\subsection{Training \& Evaluation Dataset}
The experiments are conducted using the ProGAN \cite{b13} dataset for training and the ForenSynths dataset for evaluation introduced in \cite{b2}. The ProGAN dataset includes 720K training images and 4K validation images, evenly split between real and synthetic images across 20 LSUN \cite{b14} categories. The ForenSynths dataset contains synthetic images generated by 11 different models and is combined with StyleGAN2 \cite{b15} and Which face is real \cite{b16}, which are also used in the same study. Since our training dataset consists of GAN-generated images, our evaluation is also conducted on such images.

\subsection{Evaluation Scenarios}
 The evaluation methodology simulates real-world conditions through three scenarios: an ideal scenario with no perturbations, images with individual perturbations applied (JPEG compression, Gaussian blur, Gaussian noise) and images with combined application of the above three perturbations simultaneously to test model performance in edge cases.
 
\subsection{Training Details}
The ResNet-50 \cite{b17} convolutional neural network was selected as the base model, pre-trained on ImageNet \cite{b18}, with the final fully connected layer replaced for binary classification. The training process is conducted over 25 epochs with a batch size of 64, using the Adam optimizer with parameters $\beta_{1} = 0.9$ and $\beta_{2} = 0.999$. The initial learning rate is set to 0.001 and adjusted using a StepLR scheduler. Early stopping is implemented, halting training after 7 consecutive epochs without improvement or when the learning rate reaches $10^{-6}$. The primary loss function used is Binary Cross-Entropy (BCE) Loss and for the feature similarity of perturbed and original images, Mean Squared Error (MSE) Loss or Cosine Similarity Loss are utilized alongside the BCE Loss.
    
\subsection{Pool of Image Augmentations}
The augmentation strategies used in this study are selected to cover a wide range of image transformations, applied either individually or combined. The augmentations include Random Horizontal Flipping, Random Cropping, JPEG Compression (QF = [30,100]), Gaussian Blurring ($\sigma$ = [0,3] from a uniform distribution), Gaussian Noise Injection ($\sigma$ = [0,2] from a uniform distribution), Sharpening ($s_{factor} = 2$), Contrast adjustment, Color Jitter (with parameters for brightness, contrast, saturation, and hue), Grayscale conversion, Color Inversion, AutoAugment \cite{b19}, RandAugment \cite{b20}, and Resizing to $224\times224$ or $128\times128$ with or without prior cropping. The first two augmentations—Random Horizontal Flipping and Random Cropping—are applied consistently to all images and models, as they are widely recognized as best practices. Only in the case of resizing to $224\times224$, random cropping is not applied. The remaining augmentations are applied with a probability of $P = 0.5$ to each image sample.

The specific augmentations are selected for two main reasons: to cover a broad range of transformations (geometric distortions, noise, color changes, compression) for better model generalization and to align with the best practices.

\subsection{Genetic Algorithm Details}
In the case of the Genetic Algorithm, training is performed on a subset of four ProGAN categories (car, cat, chair, horse) for 23 epochs, using the mean Average Precision (mAP) score under the combined perturbation scenario (JPEG, Gaussian Blur, Gaussian Noise) as the fitness function. The algorithm explores augmentation combinations across 8 generations, with a population size of 5 solutions per generation, applying single-point crossover, 10\% mutation probability, and tournament selection. Due to computational constraints and time limitations, we experiment with a relatively small number of generations to balance efficiency and performance. The PyGAD \cite{b21} library is used for the implementation.

\section{Experimental Results}

\subsection{Key findings}
\begin{figure*}[htbp]
\centerline{\includegraphics[clip, trim=0cm 3.2cm 0cm 1.1cm, width=\textwidth]{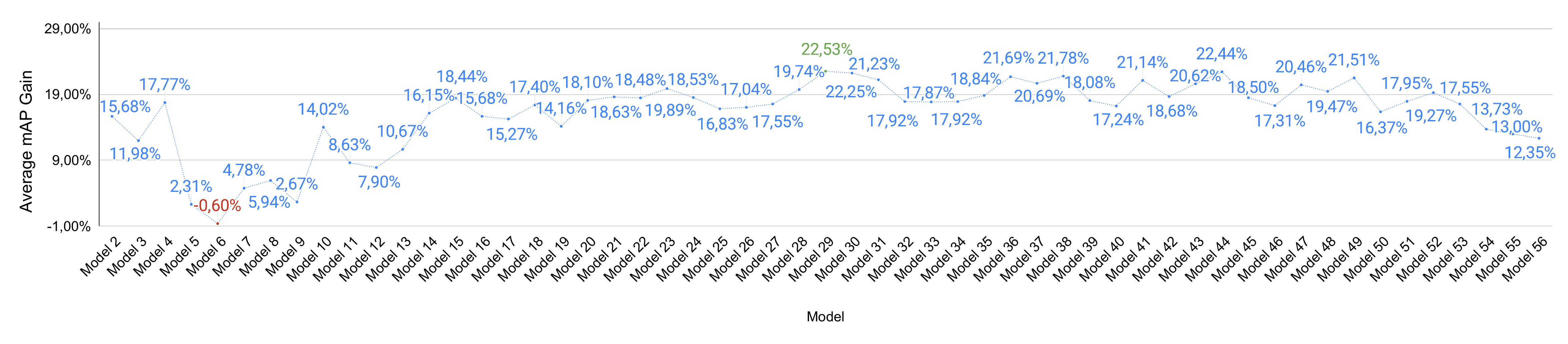}}
\caption{Average mAP Gain models trained with different augmentations following the greedy approach. \textit{Model 1 is the baseline model trained without augmentations, therefore it is not included in the figure.}}
\label{fig:avg-map-gain}
\end{figure*}

Our results indicate that augmentation techniques significantly enhance model performance. 
Models trained with augmentations selected via the Greedy approach achieve positive Average mAP Gain as demonstrated in Fig. \ref{fig:avg-map-gain}. The combination of JPEG compression, Gaussian Blur, and Color Invert results in the highest mAP Gain of 22.53\%. Models trained with AutoAugment reach a mAP of 96.15\%, outperforming the best solution of \cite{b2} in the evaluation scenario of unperturbed images, but reduce performance when evaluated on images with perturbations. These outcomes underscore that the optimal augmentation strategy boosts baseline performance and confers significant resilience to real-world distortions.

\subsection{Effect of number \& type of augmentations}
\begin{figure}[htbp]
\centerline{\includegraphics[clip, trim=0cm 0.5cm 1cm 0.5cm, width=0.5\textwidth]{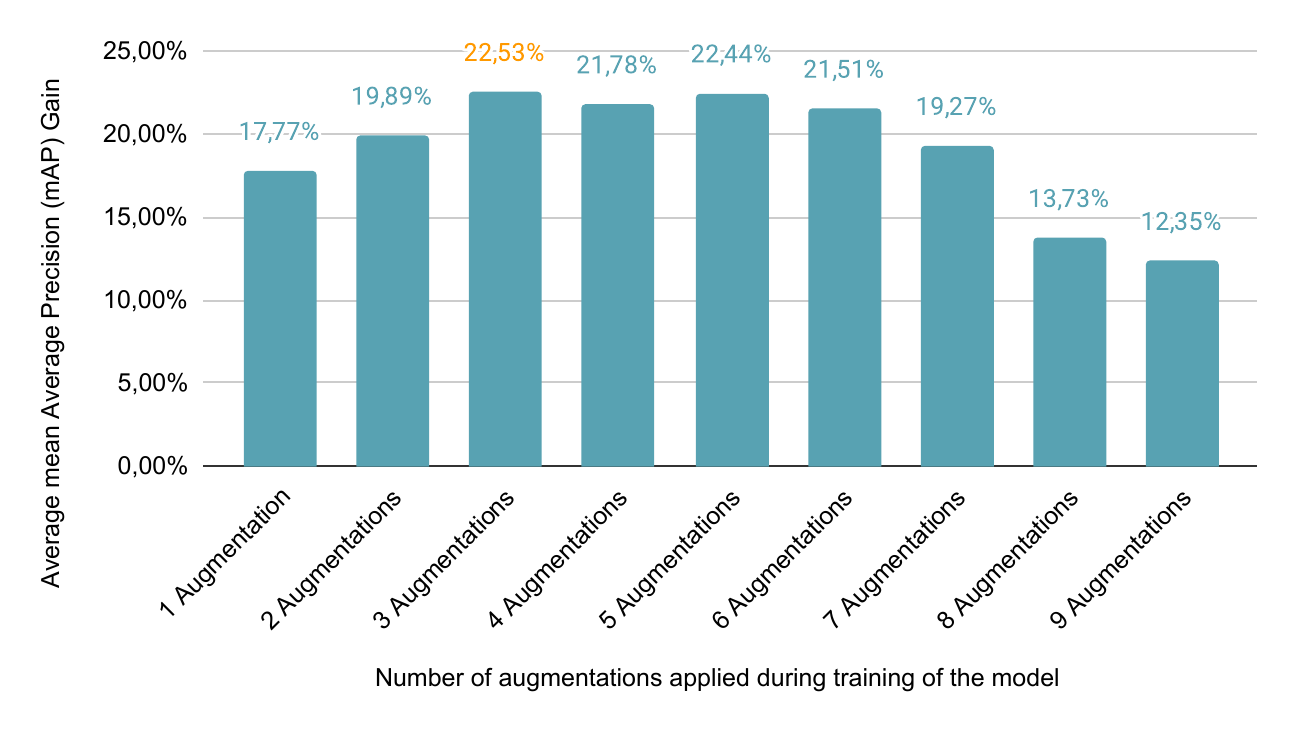}}
\caption{Average mAP Gain of the best model in each category of number of augmentations applied during training, as derived using the greedy approach.}
\label{fig:avg-map-gain-best}
\end{figure}

The results obtained using the greedy approach indicate that using more than three augmentations during training does not improve model performance and may even reduce effectiveness, likely due to alterations in original image features. Fig. \ref{fig:avg-map-gain-best} shows the Average mAP Gain of the best models per group based on the number of augmentations used. Additionally, color-space augmentations have minimal impact, suggesting that structural and textural features are more critical.

\subsection{Robustness to perturbations}
\begin{figure}[htbp]
\centerline{\includegraphics[clip, trim=0cm 0.4cm 1.5cm 0.5cm, width=0.45\textwidth]{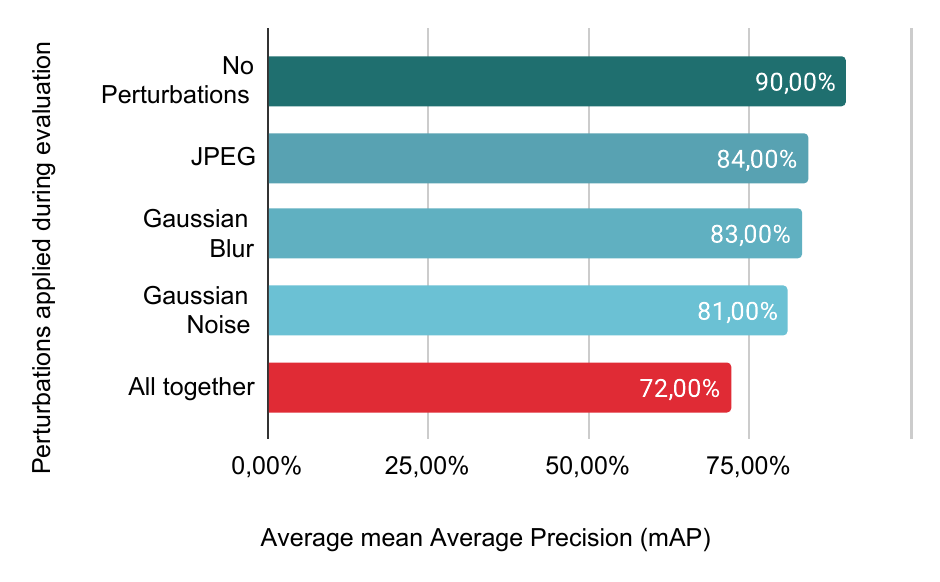}}
\caption{Average mAP of all models in each evaluation scenario as derived using the greedy approach.}
\label{fig:avg-map-per-eval-setting}
\end{figure}

\begin{figure}[htbp]
\centerline{\includegraphics[clip, trim=0cm 1.2cm 0cm 1.5cm, width=0.5\textwidth]{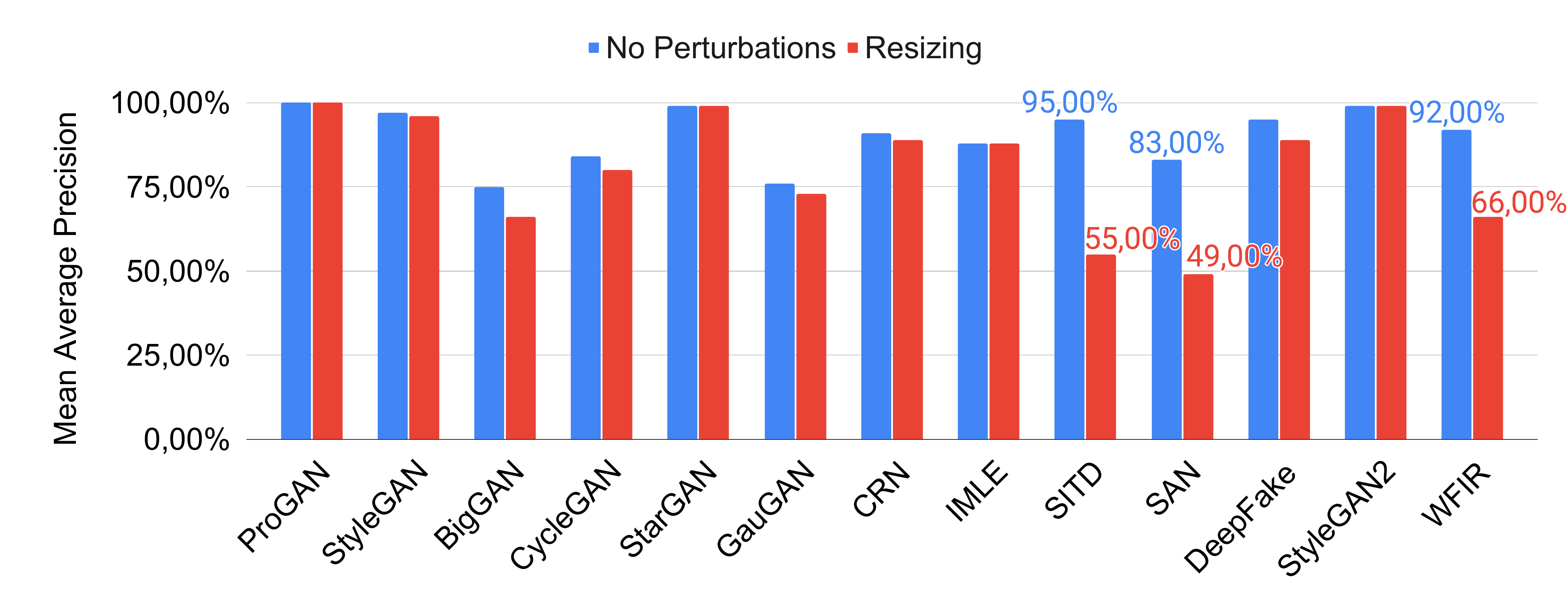}}
\caption{The mean Average Precision of models evaluated on resized images presented per dataset, as derived using the greedy approach.}
\label{fig:map-resizing}
\end{figure}

Fig. \ref{fig:avg-map-per-eval-setting} presents the Average mAP of all models from the greedy approach per evaluation scenario. The results confirm that perturbations, especially when combined together, challenge synthetic image detectors. During evaluation with combined perturbations, we see an absolute drop of 6\% - 18\% in the average mAP of models, compared with the scenario of unperturbed images. In addition, we evaluate models on images that have undergone resizing to $224 \times 224$. As expected, this is the most challenging scenario. The impact of resizing is obvious mainly on the datasets with images of higher resolution like Seeing in the Dark \cite{b22}, Second-Order Attention Network \cite{b23}, Which face is real \cite{b16}, with a relevant drop in the average mAP of all our models in the range of 28.26\% - 42.11\%. Fig. \ref{fig:map-resizing} presents the results per evaluation dataset.

\subsection{Genetic algorithm approach}
The genetic algorithm quickly identified effective augmentations, with JPEG, Gaussian Noise, Sharpen, and Contrast achieving an mAP of 71.42\% in the first generation. By generation 6, the mAP slightly improves to 71.53\%. However, the genetic algorithm requires significantly more computational time and resources, and its performance gains are marginal compared to the best model by the greedy approach (mAP = 78.31\%). Given the limitations the full potential of the genetic algorithm may not have been realized.

\subsection{Dual-criteria optimization approach}
\begin{figure}[htbp]
\centerline{\includegraphics[clip, trim=0cm 0.8cm 0cm 0.5cm, width=0.5\textwidth]{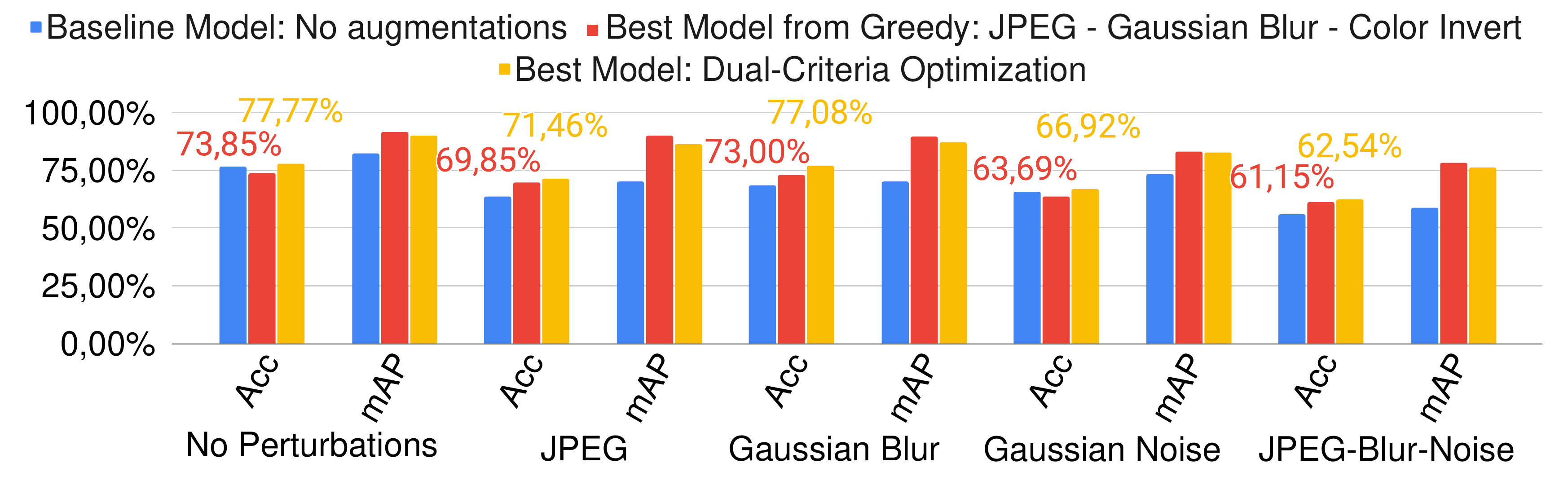}}
\caption{Comparison of: a) baseline model trained without augmentations, b) model trained with the best combination of augmentations from the greedy approach and c) best model trained with the dual-criteria optimization process with $\lambda_{1}=1, \lambda_{2}=0.2$ and the Cosine Loss function.}
\label{fig:dual-criteria-results}
\end{figure}

The dual-criteria approach, incorporating Cosine Loss alongside the standard classification loss, does not lead to further increase in mAP, compared with the best model from the greedy approach but results in a 1.39\% - 4.08\% absolute increase in accuracy. This suggests that while the additional criterion does not enhance the model’s overall ability to distinguish synthetic from real images, it contributes to more stable and reliable predictions. This is particularly important in real-world applications, where ensuring reliable classification under varying conditions is crucial. In Fig. \ref{fig:dual-criteria-results}, we present the evaluation results from the baseline model trained without augmentations, the best model from the greedy approach, as well as the best model using this approach.

\section{Conclusion and Future Considerations}
This research presents a comprehensive approach to improve synthetic image detection on perturbed images. Our findings confirm the importance of carefully selecting augmentations to enhance model performance under perturbations. By exploring augmentation techniques and introducing dual-criteria optimization, the study shows significant improvements in model performance under various scenarios. The greedy algorithm proves effective in identifying an efficient augmentation combination, with the best-performing set consisting of JPEG Compression, Gaussian Blur, and Color Invert. The genetic algorithm finds reasonable combinations, suggesting that more compute could yield better results. The dual-criteria optimization approach reveals potential for improving model stability under perturbations and remains open for investigation.

Limitations were observed, as augmentations effectiveness varied across datasets, with some techniques reducing performance on high-resolution images. Future work should focus on further refining augmentation strategies, exploring additional loss functions, and testing models in the wild.
While our methodology improves robustness under real-world perturbations, its applicability on diffusion-generated images remains an open question. Diffusion-based images exhibit different frequency traces than GAN-based ones \cite{b7}, and applying our pipeline may reveal different optimal augmentations, a direction for future work. Finally, developing real-time tools like browser extensions could improve model applicability.

\section*{Acknowledgment}
This work has been partially funded by the Horizon Europe project AI4Trust (GA No. 101070190). The results presented have been produced using the Aristotle University of Thessaloniki (AUTh) High Performance Computing Infrastructure and Resources.


\begin{thebibliography}{00}

\bibitem{b1} L. Chai, D. Bau, S.-N. Lim, and P. Isola, “What Makes Fake Images Detectable? Understanding Properties that Generalize.,” in ECCV (26), 2020, vol. 12371, pp. 103–120 [Online]. Available: http://dblp.uni-trier.de/db/conf/eccv/eccv2020-26.html\#ChaiBLI20


\bibitem{b2} S.-Y. Wang, O. Wang, R. Zhang, A. Owens, and A. A. Efros, “CNN-Generated Images Are Surprisingly Easy to Spot... for Now.,” in CVPR, 2020, pp. 8692–8701 [Online]. Available: http://dblp.uni-trier.de/db/conf/cvpr/cvpr2020.html\#WangW0OE20

\bibitem{b3} U. Ojha, Y. Li, and Y. J. Lee, “Towards Universal Fake Image Detectors that Generalize Across Generative Models.,” in CVPR, 2023, pp. 24480–24489 [Online]. Available: http://dblp.uni-trier.de/db/conf/cvpr/cvpr2023.html\#OjhaLL23

\bibitem{b4} H. Liu, Z. Tan, C. Tan, Y. Wei, J. Wang, and Y. Zhao, “Forgery-aware Adaptive Transformer for Generalizable Synthetic Image Detection.,” in CVPR, 2024, pp. 10770–10780 [Online]. Available: http://dblp.uni-trier.de/db/conf/cvpr/cvpr2024.html\#LiuTTW0Z24

\bibitem{b5} C. Koutlis and S. Papadopoulos, “Leveraging Representations from Intermediate Encoder-Blocks for Synthetic Image Detection.,” in ECCV (72), 2024, vol. 15130, pp. 394–411 [Online]. Available: http://dblp.uni-trier.de/db/conf/eccv/eccv2024-72.html\#KoutlisP24

\bibitem{b6} A. Radford et al., "Learning Transferable Visual Models From Natural Language Supervision," in Proceedings of the 38th International Conference on Machine Learning (ICML), Virtual Event, 18–24 July 2021, vol. 139, Proceedings of Machine Learning Research, pp. 8748–8763, 2021. [Online]. Available: http://proceedings.mlr.press/v139/radford21a.html.

\bibitem{b7} R. Corvi, D. Cozzolino, G. Zingarini, G. Poggi, K. Nagano, and L. Verdoliva, “On The Detection of Synthetic Images Generated by Diffusion Models.,” in ICASSP, 2023, pp. 1–5 [Online]. Available: http://dblp.uni-trier.de/db/conf/icassp/icassp2023.html\#CorviCZPNV23

\bibitem{b8} D. Gragnaniello, D. Cozzolino, F. Marra, G. Poggi, and L. Verdoliva, “Are GAN Generated Images Easy to Detect? A Critical Analysis of the State-Of-The-Art.,” in ICME, 2021, pp. 1–6 [Online]. Available: http://dblp.uni-trier.de/db/conf/icmcs/icme2021.html\#GragnanielloCMP21

\bibitem{b9} U. Ojha, Y. Li, and Y. J. Lee, “Towards Universal Fake Image Detectors that Generalize Across Generative Models.,” in CVPR, 2023, pp. 24480–24489 [Online]. Available: http://dblp.uni-trier.de/db/conf/cvpr/cvpr2023.html\#OjhaLL23

\bibitem{b10} B. Chen, J. Zeng, J. Yang, and R. Yang, ``DRCT: Diffusion Reconstruction Contrastive Training towards Universal Detection of Diffusion Generated Images,'' in \textit{Proc. 41st Int. Conf. Mach. Learn. (ICML)}, 2024. [Online]. Available: https://openreview.net/forum?id=oRLwyayrh1

\bibitem{b11} D. Cozzolino, G. Poggi, R. Corvi, M. Nießner, and L. Verdoliva, “Raising the Bar of AI-generated Image Detection with CLIP.,” in CVPR Workshops, 2024, pp. 4356–4366 [Online]. Available: http://dblp.uni-trier.de/db/conf/cvpr/cvprw2024.html\#CozzolinoPCNV22

\bibitem{b12} M. Schinas and S. Papadopoulos, “SIDBench: A Python framework for reliably assessing synthetic image detection methods.,” in MAD@ICMR, 2024, pp. 55–64 [Online]. Available: http://dblp.uni-trier.de/db/conf/mir/mad2024.html\#SchinasP24

\bibitem{b13} T. Karras, T. Aila, S. Laine, and J. Lehtinen, “Progressive Growing of GANs for Improved Quality, Stability, and Variation.,” in ICLR, 2018 [Online]. Available: http://dblp.uni-trier.de/db/conf/iclr/iclr2018.html\#KarrasALL18

\bibitem{b14} F. Yu, A. Seff, Y. Zhang, S. Song, T. Funkhouser, and J. Xiao, ``LSUN: Construction of a large-scale image dataset using deep learning with humans in the loop,'' arXiv preprint arXiv:1506.03365, 2016. [Online]. Available: https://arxiv.org/abs/1506.03365

\bibitem{b15} T. Karras, S. Laine, M. Aittala, J. Hellsten, J. Lehtinen and T. Aila, "Analyzing and Improving the Image Quality of StyleGAN," 2020 IEEE/CVF Conference on Computer Vision and Pattern Recognition (CVPR), Seattle, WA, USA, 2020, pp. 8107-8116, doi: 10.1109/CVPR42600.2020.00813.

\bibitem{b16} ``Which face is real?,'' [Online]. Available www.whichfaceisreal.com

\bibitem{b17} K. He, X. Zhang, S. Ren and J. Sun, "Deep Residual Learning for Image Recognition," 2016 IEEE Conference on Computer Vision and Pattern Recognition (CVPR), Las Vegas, NV, USA, 2016, pp. 770-778, doi: 10.1109/CVPR.2016.90.

\bibitem{b18} O. Russakovsky, J. Deng, H. Su, J. Krause, S. Satheesh, S. Ma, Z. Huang, A. Karpathy, A. Khosla, M. Bernstein, A. C. Berg, and L. Fei-Fei, "ImageNet Large Scale Visual Recognition Challenge," International Journal of Computer Vision, vol. 115, no. 3, pp. 211–252, Dec. 2015. [Online]. Available: https://doi.org/10.1007/s11263-015-0816-y.

\bibitem{b19} E. D. Cubuk, B. Zoph, D. Mané, V. Vasudevan and Q. V. Le, "AutoAugment: Learning Augmentation Strategies From Data," 2019 IEEE/CVF Conference on Computer Vision and Pattern Recognition (CVPR), Long Beach, CA, USA, 2019, pp. 113-123, doi: 10.1109/CVPR.2019.00020. keywords: {Deep Learning},


\bibitem{b20} E. D. Cubuk, B. Zoph, J. Shlens, and Q. Le, “RandAugment: Practical Automated Data Augmentation with a Reduced Search Space.,” in NeurIPS, 2020 [Online]. Available: http://dblp.uni-trier.de/db/conf/nips/neurips2020.html\#CubukZS020

\bibitem{b21} Gad, A.F. PyGAD: an intuitive genetic algorithm Python library. Multimed Tools Appl 83, 58029–58042 (2024). https://doi.org/10.1007/s11042-023-17167-y

\bibitem{b22} C. Chen, Q. Chen, J. Xu, and V. Koltun, “Learning to See in the Dark.,” in CVPR, 2018, pp. 3291–3300 [Online]. Available: http://dblp.uni-trier.de/db/conf/cvpr/cvpr2018.html\#ChenCXK18

\bibitem{b23}T. Dai, J. Cai, Y. Zhang, S. -T. Xia and L. Zhang, "Second-Order Attention Network for Single Image Super-Resolution," 2019 IEEE/CVF Conference on Computer Vision and Pattern Recognition (CVPR), Long Beach, CA, USA, 2019, pp. 11057-11066, doi: 10.1109/CVPR.2019.01132.
keywords: {Low-level Vision;Deep Learning}

\bibitem{b24}N. Purnekar, A. Kaul, T. Subramanian and M. Barni, "Improving the Robustness of Synthetic Images Detection by Means of Print and Scan Augmentation," In Proceedings of the 2024 ACM Workshop on Information Hiding and Multimedia Security (IH\&MMSec), New York, NY, USA, 65–73. https://doi.org/10.1145/3658664.3659635 


\end{thebibliography}
\end{document}